\documentclass[letterpaper, 10 pt, conference]{ieeeconf}  

\IEEEoverridecommandlockouts                              

\usepackage{graphicx}
\usepackage{multirow}
\usepackage{amssymb}
\usepackage{amsmath}
\usepackage{amstext}
\usepackage{comment}
\usepackage[dvipsnames]{xcolor}
\usepackage{pifont}
\usepackage{array}
\newcommand{\PreserveBackslash}[1]{\let\temp=\\#1\let\\=\temp}
\newcolumntype{C}[1]{>{\PreserveBackslash\centering}p{#1}}
\newcolumntype{R}[1]{>{\PreserveBackslash\raggedleft}p{#1}}
\newcolumntype{L}[1]{>{\PreserveBackslash\raggedright}p{#1}}
\usepackage[colorlinks=true,citecolor=blue,urlcolor=blue]{hyperref}
\usepackage{cite}
\usepackage{textcomp}

\title{\LARGE \bf
Relational Graph Learning on Visual and Kinematics Embeddings for Accurate Gesture Recognition in Robotic Surgery
}

\author{
Yonghao Long, Jie Ying Wu, Bo Lu, Yueming Jin, Mathias Unberath, \\ Yun-Hui Liu, Pheng Ann Heng and Qi Dou
\thanks{This project was supported by CUHK Shun Hing Institute of Advanced Engineering (project MMT-p5-20), CUHK T Stone Robotics Institute, Hong Kong RGC TRS Project No.T42-409/18-R, and Multi-Scale Medical Robotics Center InnoHK under grant 8312051.}
\thanks{Y. Long, Y. Jin, P. A. Heng and Q. Dou are with the Department of Computer Science and Engineering, The Chinese University of Hong Kong. P. A. Heng and Q. Dou are also with the T Stone Robotics Institute, CUHK. B. Lu and Y. H. Liu are with the T Stone Robotics Institute, the Department of Mechanical and Automation Engineering, The Chinese University of Hong Kong. J. Y. Wu and M. Unberath are with the Department of Computer Science, Johns Hopkins University.}
\thanks{\textit{Corresponding author: Qi Dou (qidou@cuhk.edu.hk).}}%
}
\begin{document}
\maketitle
\thispagestyle{empty}
\pagestyle{empty}
\begin{abstract}

Automatic surgical gesture recognition is fundamentally important to enable intelligent cognitive assistance in robotic surgery. With recent advancement in robot-assisted minimally invasive surgery, rich information including surgical videos and robotic kinematics can be recorded, which provide complementary knowledge for understanding surgical gestures. However, existing methods either solely adopt uni-modal data or directly concatenate multi-modal representations, which can not sufficiently exploit the informative correlations inherent in visual and kinematics data to boost gesture recognition accuracies. In this regard, we propose a novel online approach of \textbf{m}ulti-modal \textbf{r}elational \textbf{g}raph \textbf{net}work (\emph{i.e., MRG-Net}) to dynamically integrate visual and kinematics information through interactive message propagation in the latent feature space. In specific, we first extract embeddings from video and kinematics sequences with temporal convolutional networks and LSTM units. Next, we identify multi-relations in these multi-modal embeddings and leverage them through a hierarchical relational graph learning module. The effectiveness of our method is demonstrated with state-of-the-art results on the public JIGSAWS dataset, outperforming current uni-modal and multi-modal methods on both suturing and knot typing tasks. Furthermore, we validated our method on in-house visual-kinematics datasets collected with da Vinci Research Kit (dVRK) platforms in two centers, with consistent promising performance achieved. Our code and data are released at: \href{https://www.cse.cuhk.edu.hk/\%7eyhlong/mrgnet.html}{https://www.cse.cuhk.edu.hk/\texttildelow yhlong/mrgnet.html}.

\end{abstract}

\section{Introduction}

Robot-assisted surgery, with a short but remarkable chronicle~\cite{moustris2011evolution}, has dramatically extended the dexterity and overall capability of surgeons, and plays an important role in modern minimally invasive surgery.
Robotic systems enable precise control, efficient manipulation and vivid observation for the surgical procedures, yielding rich sources of information~\cite{guthart2000intuitive}.
Intelligent understanding of such complex surgical procedure is highly desired for facilitating cognitive assistance. To this end, automatic gesture recognition is fundamentally required for supporting higher-level perception such as surgical decision making~\cite{maier2017surgical}, surgical skill assessment~\cite{poursartip2018analysis} and surgical task automation~\cite{nagy2019dvrk} towards the next generation of operating theatres.
However, accurately recognizing on-going surgical gesture is challenging, due to the complex multi-step actions, frequent state transitions, disturbance in sensor data, various manipulation habits and proficiency of different surgeons.

To address above challenges in automatic surgical gesture recognition, a set of methods have been developed in the past decade. Some methods were based on processing sequential robotic kinematics data (e.g., the position and velocity of the tool tips), using traditional machine learning methods such as variants of hidden Markov models~\cite{tao2012sparse,varadarajan2009data}, linear classifiers with hand-crafted metrics~\cite{zappella2013surgical} and recent deep learning methods such as long short-term memory (LSTM)~\cite{dipietro2016recognizing}, multi-task recurrent neural network~\cite{van2020multi} and multi-scale recurrent network (offline)~\cite{gurcan2019surgical}.
In the meanwhile, purely video based solutions have been intensively explored in the recent years, employing deep convolutional neural networks for extracting high-quality visual features.
Promising gesture recognition results have been achieved relying on temporal convolutional network (TCN)~\cite{lea2017temporal}, recurrent convolutional network~\cite{jin2017sv}, 3D convolutional network~\cite{funke2019using} and symmetric dilated convolution (offline)~\cite{zhang2020symmetric} to extract representative visual features.
However, all these solutions only adopted a single source of information,
without considering the multi-modal joint knowledge inherent in kinematics and visual data
synchronously recorded in robotic systems.

\begin{figure*}[t]
  \centering
  \includegraphics[width=0.93\textwidth]{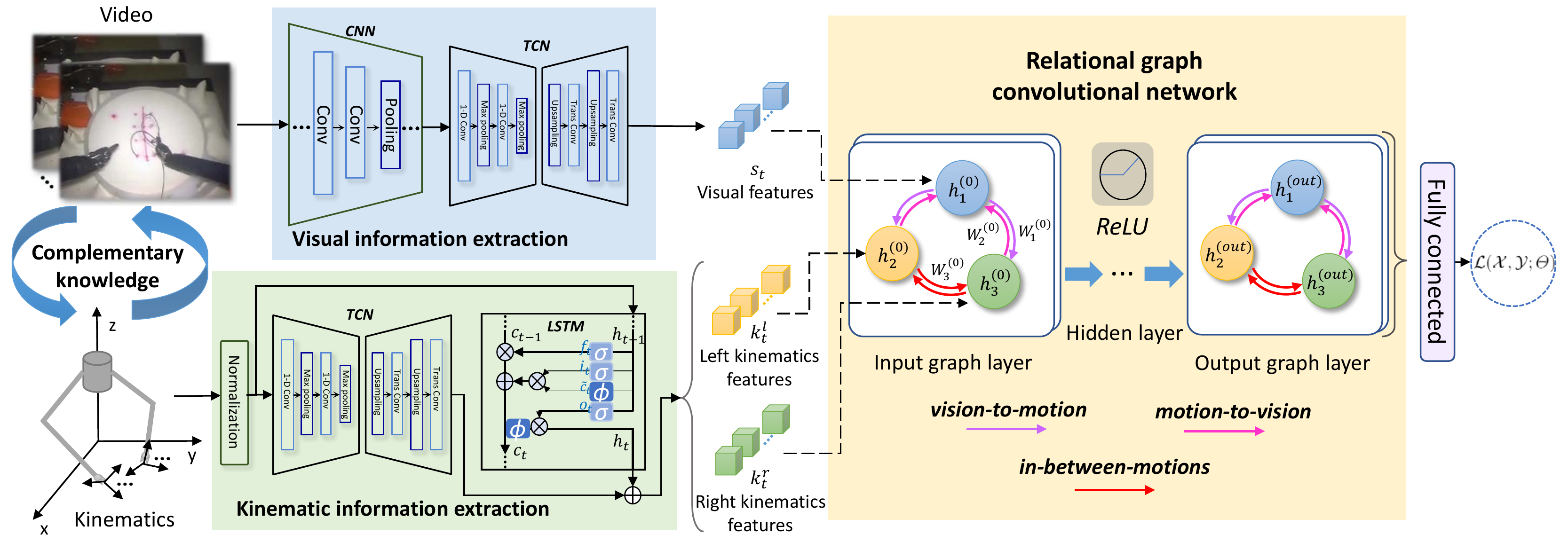}
  \vspace{-5mm}
  \caption{The overview of our proposed multi-modal relational graph network for surgical gesture recognition in robot-assisted surgery.}
  \label{fig1}
  \vspace{-6mm}
\end{figure*}

As we understand, the kinematics and video data can be regarded as the hands and eye of the surgical robot, with eye giving visual guidance information for two hands collaboratively conducting specific actions, while hands drive changes in the visual scene.
In this regard, there are complementary information and joint knowledge contained in the kinematics and video data which are crucial to help gesture recognition.
Several recent works have attempted to develop multi-modal learning methods.
For instance, some unsupervised multi-modal methods have been proposed to handle the problem of time-consuming annotation~\cite{murali2016tsc,zhao2018fast}.
Lea \textit{et al.}~\cite{lea2016learning} designed a latent convolutional skip-chain conditional random field model with variables of scene-based features and kinematics data.
The work of Fusion-KV~\cite{qin2020temporal} learned individual networks for each modality, and combined their predictions through weighted voting at an output level. Qin \textit{et al.}~\cite{qin2020davincinet} further improved Fusion-KV with an attention-based LSTM decoder to predict the surgical state using concatenated multi-modal features.
Despite gaining performance improvement, these multi-modal feature fusions seem straightforward. How to dynamically integrate the multiple sources of information in latent feature space, to reveal and leverage the underlying relationships inherent in kinematics sequences and video scenes, is important yet still remains underexplored.

Recently, graph neural networks have been increasingly receiving research interest, due to their capability to model non-Euclidean relationships among entities~\cite{dwivedi2020benchmarkgnns,scarselli2008graph,zhou2018graph}. Graph convolutional networks (GCN)~\cite{kipf2016semi} have widely demonstrated promising performances on applications in various domains including image classification~\cite{wang2018zero}, neural machine translation~\cite{marcheggiani2018exploiting}, social relationship understanding~\cite{wang2018deep}, etc.
Specifically for robotic surgery related scenarios, there have been pilot studies applying graph neural networks for tool detection in surgical videos~\cite{wang2019graph},
3D point cloud classification~\cite{weibel2019robust,weibel2019addressing} and surgical activity recognition from robotic joint pose estimation~\cite{sarikaya2020towards}.
These achievements inspired us to explore the potential of graph learning for modeling distinct multi-modal data recorded in robotic surgery.

In this paper, we propose a novel \textbf{m}ulti-modal \textbf{r}elational \textbf{g}raph \textbf{net}work (i.e., MRG-Net) to effectively exploit important yet complex relationships in robotics visual and kinematics data for accurate surgical gesture recognition.
Specifically, we first extract the high-level embeddings from video scenes and kinematics sequences with temporal convolutional networks and LSTM units. Then, we leverage relational graph convolutional network to incorporate complementary sources of information and model the underlying multiple types of relations.
Our main contributions are summarized as follows:

\begin{itemize}
	\item We, for the first time, propose a novel online relational graph learning based framework to exploit the joint information with useful relationships in video and kinematics data for accurate surgical gesture recognition.
	\item We evaluated our proposed method on the public robotic surgery dataset JIGSAWS, and set new state-of-the-art \\
	results on both suturing and knot typing tasks, showing the efficacy of combined usage of visual and kinematics information for robotic intelligence.
	\item We have extensively validated our method on in-house datasets collected from da Vinci Research Kit (dVRK) platforms in two centers (i.e., CUHK and JHU) with consistent promising results achieved, demonstrating the general effectiveness of our proposed method.
\end{itemize}

\section{Methods}

In robot-assisted surgery, the robotic system can generate video frames from endoscopy and kinematics sequences from multiple robotic arms, which are later synchronized to the video timestamps. The overview of our proposed network is shown in Fig.~\ref{fig1}.
Our network consists of three components, i.e. visual and kinematic information extraction modules, as well as the relational graph convolutional network. We first extract the visual and kinematic embeddings with visual and kinematic information extraction modules, and then model the complementary information and integrate the informative joint knowledge of these multi-modal features with relational graph convolutional network. As a whole, MRG-Net forms a multi-input single-output design to predict the probability distributions of surgical gestures at each time step.


\vspace{-1mm}
\subsection{Visual and Kinematic Embeddings Extraction}

The first part of the network is the visual and kinematic information extraction modules which extract representative descriptors from each of the following streams respectively: the video frames and the kinematics sequences of robotic left and right arms.
Regarding the visual information, for each time step $t$, current video frames (RGB image) $I_t$ is forwarded to a standard CNN backbone (in this case we leverage a 18-layer deep residual network (ResNet-18)~\cite{he2016deep}), yielding a vector of spatial feature ${u}_t$. For the entire video sample, the series of $\{u_t\}_{t=1}^T$ are input to a temporal convolution module, which adopts an encoder-decoder operation to hierarchically capture relationships across frames at multi-time-scales, yielding stronger spatio-temporal video features of $\{s_t\}_{t=1}^T$.

For the kinematics data, our feature extractor incorporates TCN and LSTM in parallel for modeling the complex sequential information of physical elements and for better capturing the local and longer-term temporal dependencies.
Specifically, the input to TCN stacks the kinematics variables from all time steps, followed by temporal convolutions, pooling, channel-wise normalization and upsampling to encode kinematics features as $\{k_t^{tcn}\}^T_{t=1}$. Meanwhile, LSTM obtains the feature $k^{lstm}_t$ of the current step, by inputting a sequence of kinematics of all its previous steps, for capturing the long-term dependencies in motions. Then, the $k_t^{tcn}$ and $k_t^{lstm}$ are averaged to represent the kinematics feature as $k_t$. Note that we separately encode left and right kinematics as $\{k_t^{l},k_t^{r}\}_{t=1}^T$, since the two robotic arms may conduct different actions and serve for specific purposes in surgery.

\subsection{Fusion of Multi-modal Embeddings with Graph Learning}

Next, a designed graph learning module is subsequently adopted to fuse the above extracted high-level embeddings $\{s_t, k_t^{l}, k_t^{r}\}$ of each time step.
These features have already gained temporal information within each source of time-series data. The following key issue is to effectively exploit their joint knowledge by imposing structured interactions between multi-modal features for accurate gesture recognition.
Intuitively, the graph learning layer plays a role for differentiable message passing framework~\cite{gilmer2017neural}.
Specifically, we denote our graph as $G \! = \! \{\mathcal{V},\mathcal{E}, \mathcal{R}\}$ with nodes $v_i \! \in \! \mathcal{V}$ and edges  $(v_i, r, v_j) \in \mathcal{E}$ where $r\in \mathcal{R}$ is a relation. As shown in Fig.~\ref{fig1}, there are three node entities corresponding to video, left kinematics and right kinematics, whose associated feature descriptors of $\{h_1,h_2,h_3\}$ are initialized as $\{s_t, k_t^{l}, k_t^{r}\}$. These descriptors are then updated by aggregating messages from neighboring nodes with a parameterized propagation rule, which can be generally written as:
\begin{equation}
h_i^{(l+1)} = \sigma\left( \sum_{j\in \mathcal{N}_i} f_m(h_i^{(l)}, h_j^{(l)})\right),\\
\end{equation}
where $h_i^{(l)}$ is the hidden state of node $v_i$ in the $l$-th graph network layer, $\mathcal{N}_i$ is the set of indices of all nodes which are connected with node $i$, the $f_m(\cdot, \cdot)$ denotes the function for accumulating incoming messages from a relational neighbor, and $\sigma(\cdot)$ is the element-wise non-linear activation, i.e., the ReLU in our model.

Intuitively, such interactive feature fusion is important to impose a learnable message passing process for the nodes which have some relations with each other.
Given that the multiple data sources from robotic surgery contain plenty of complementary information, effectively digging the inherent useful relationships among them is difficult while critical to boost the performance of gesture recognition.

\subsection{Multi-relation Modelling in Graph Latent Space}

In our scenario of robot-assisted surgery, there are at least three important types of relations in the video and kinematics data. Specifically, the first is the \emph{vision-to-motion} relation which can be understood as the human's perception with the ``eyes" to provide guidance information for the ``hands" to move. Inversely, the second relation is \emph{motion-to-vision} which reflects the mechanism of ``hands" giving feedback to ``eyes" and also resembles hand-eye coordinates projection of robotic vision~\cite{tsai1989new}. The last relation is \emph{in-between-motions} of left and right arms, which can be considered as two ``hand" assisting each other to complete a task.
The widely-used conventional graph convolutional network~\cite{kipf2016semi} is inadequate to handle various different types of relations, given that its undirected graph with $|\mathcal{R}| \! = \! 1$ is insufficient to model multi-relations of nodes. Instead, we leverage the more powerful relational graph learning scheme,
so that our $G$ is a directed graph endowing a higher capacity for modeling multiple types of directed edges between nodes. In this way, the parameterized propagation function $f_m(\cdot,\cdot)$ in Eq.(1) becomes relation-specific, where the forwarding message update to node $i$ from a relational node $j$ is elaborated as $c_{i,r}  h_j^{(l)}W_r^{(l)}$. The $W_r^{(l)}$ represents a trainable transformation matrix that is uniquely associated to one certain type of relation $r \in \mathcal{R}$.
In other words, different relation types use individual matrices, and only directed edges of the same relation type share their weights.
The parameter $c_{i,r}$ is a normalization constant that correlates to the structure of the graph.
In this way, the layer-wise propagation is achieved by accumulating the message updates through a normalized sum of all neighbor nodes under all relation types.

Hierarchically, we stack two such relational graph learning layers, with each single layer having its separate set of projection weights $\{W_r^{(l)}\}_{l=0}^1$. No deeper layers are added to alleviate the over-smooth problem~\cite{li2018deeper} of GCN, and our preliminary experiments also evidenced that additional layers yielded worse performance yet with heavier computations.
After feature interactions, the final output representation associated with each node is computed by:


\begin{scriptsize}
\begin{equation}
h_i^{(\text{out})} = \sigma\left(\sum_{r\in \mathcal{R}}\sum_{j\in N_i^r} {c_{i,r}}
\sigma\left(\sum_{r\in \mathcal{R}}\sum_{j\in N_i^r}{c_{i,r}}h_j^{(0)} W_r^{(0)}\right) W_r^{(1)}\right),
\label{rgcn2}
\end{equation}
\end{scriptsize}
where the $\mathcal{N}_i^r$ denotes the set of neighbor indices of node $i$ under a relation type $r \in \mathcal{R}$. Specifically in our model, we identify three different types of relations with $|\mathcal{R}|=3$.
For instance, as shown in Fig.~\ref{fig1}, the video node $h_1$ receives messages from kinematics nodes $\{h_2,h_3\}$, both under the relation type of \emph{motion-to-vision} (cf. $W_2^{(0)}$ in brown arrow).
The left kinematics node $h_2$ receives messages from video node $h_1$ under relation type of \emph{vision-to-motion} (cf. $W_1^{(0)}$ in purple arrow), and from right kinematics $h_3$ under relation type of \emph{in-between-motions} (cf. $W_3^{(0)}$ in red arrow).
The weight $c_{i, r}$ is heuristically set as $1/|\mathcal{N}_i^r|$.
We exclude the self-loop aggregation for each node, with the consideration that, for our graph classification task, the self-loop information tends to result in feature redundancy during the update process, weakening the messages propagated from neighbor nodes (i.e., other modalities). Note that such a practice does not cause knowledge leakage since the hierarchical propagation can compensate those self-contained information through iterative interactions among nodes.
In addition, the regularization strategy of basis decomposition~\cite{schlichtkrull2018modeling} is applied for $\{W_r^{(l)}\}_{l=0}^1$ to prevent rapid growth of the number of parameters with multi-relational data.

\subsection{Overall Loss Function}

After interacting the multi-modal information in latent space for capturing joint knowledge, the relational graph learning layers produce updated representations for the nodes. Recall that the hidden state for each node represents the descriptors for each modality at time step $t$, so we rephrase $\{h_i^{(\text{out})}\}_{i=1}^3$ into $\{\tilde{s}_t, \tilde{k}_t^l, \tilde{k}_t^r\}$ as the set of features for video, left and right kinematics after graph learning. They are concatenated to convey the joint knowledge, and forwarded to a fully-connected layer for obtaining the classification prediction $\hat{p}_t$ for each frame:

\begin{equation}
 \hat{p}_t=Softmax(\textbf{concat}[\tilde{s}_t, \tilde{k}_t^l, \tilde{k}_t^r] W_{\text{fc}} + b).
\end{equation}
With the situation in the robotic surgery that the duration of each conducted gesture varies widely (e.g., the gesture of ``loosening more suture" occupies only about 1\% time on average of the whole suturing task, while ``pushing needle through tissue" takes up almost 30\% task duration), we use the weighted cross-entropy loss to combat such inter-class imbalance for training samples.
Denoting $\alpha$ as the class balancing weight, $\Theta$ as MRG-Net parameters of all trainable layers, 
we optimize the overall loss function:
\begin{equation}
\mathcal{L}(\mathcal{X,Y}; \Theta) = \frac{1}{T}\sum\nolimits_{t} -\alpha \cdot \log \hat{p}_t,
\end{equation}
where $\mathcal{X}$ is the multi-modal input space, $\mathcal{Y}$ denotes the gesture categories.

\subsection{Implementation Details}
Overall, the entire framework composing of the relational graph layers and the separate video and kinematics feature extractors is trained end-to-end.
The encoder and decoder of TCN backbone consists of $3$ temporal convolutional layers with $\{64,96,128\}$ filters for encoder and $\{96,64,64\}$ filters for decoder, with the kernel size of $51$.
For the visual information, we first train the CNN backbone (ResNet-18) using video sequences, then we generate the spatial-CNN features $u_t \! \in \! \mathbb{R}^{128}$ from the pretrained backbone to train the whole model more efficiently. For kinematic data, we first convert the rotation matrix
into Euler angles,
then normalize all the data to zero mean and unit variance. The relational graph layers have 64-dimensional hidden states and output states, with dropout (rate $\!=\!0.2$) applied.

Our graph learning framework is implemented with the Deep Graph Library (DGL)~\cite{wang2019dgl} in PyTorch with an NVIDIA Titan Xp GPU. The video frames are resized to resolution 320*256 with random crop 224*224 to reduce the training parameters and prevent over-fitting.
We used the Adam~\cite{kingma2014adam} optimiser with learning rate of $5e^{-3}$ and weight decay of $5e^{-4}$ to train the proposed network. The training process took around 3 hours for a hundred epochs.
To avoid the case of coincidence, we trained the same model for three times and reported their average results.

\section{Experiments}

\subsection{Public Dataset and Evaluation Metrics}

We first extensively validate our proposed MRG-Net on the public dataset of JIGSAWS~\cite{gao2014jhu} (JHU-ISI Gesture and Skill Assessment Working Set) on two tasks (i.e., suturing and knot typing), which consist of 39 videos and 36 videos respectively alongside with kinematics data of left and right robotic arms from the \emph{da Vinci} surgical system.
The kinematics sequences include position, rotation matrix, linear velocity, rotational velocity of tool tip and angles of gripper.
A total of ten categories of surgical gestures are annotated for each single frame in suturing task and six in knot typing task (cf.~\cite{gao2014jhu} for detailed gesture definitions).
Our experimental setting adopts \emph{leave-one-user-out} cross validation, following the practice of previous works on this benchmark~\cite{ahmidi2017dataset}.
The employed evaluation metrics on JIGSAWS dataset include: i) Accuracy (\%) in the frame-wise level, which is to calculate the percentage of correctly recognized frames, ii) Edit Score~\cite{lea2016segmental}
(in range $[0,100]$, the higher score the better), which is designed to measure the performance in video segmentation level for emphasizing temporal smoothness.

\subsection{Comparison with Other State-of-the-art Methods}
\begin{table}[t]
\begin{center}
\caption{Results of different methods on JIGSAWS Suturing dataset for gesture recognition.
}\label{tab1}
\scalebox{0.94}{
\begin{tabular}{|c|c|c|c|c|}
\hline
\multirow{2}{*}{Methods} & \multicolumn{2}{|c|}{Input data} & \multirow{2}{*}{Accuracy}& \multirow{2}{*}{Edit Score} \\
\cline{2-3}
~ & ~Kin & Vid &&\\
\hline
TCN~\cite{lea2016temporal} & \checkmark  && 79.6 & 85.8\\
Forward LSTM~\cite{dipietro2016recognizing} & \checkmark  && 80.5 $\pm$ 6.2 & 75.3\\
TricorNet~\cite{ding2017tricornet} & \checkmark  & & 82.9 & 86.8\\
Bidir. LSTM~\cite{dipietro2016recognizing} & \checkmark  && 83.3 $\pm$ 5.7 & 81.1\\
Bidir GRU~\cite{dipietro2019segmenting} & \checkmark  && 84.7 $\pm$ 6.0 & 88.5\\
APc~\cite{van2020multi} & \checkmark  & & 85.5 & 85.3\\
\hline
TCN~\cite{lea2016temporal} &  & \checkmark & 81.4 & 83.1\\
Policy+Value~\cite{gao2020automatic} &  & \checkmark & 81.7 & 88.5\\
3D CNN(K)+window~\cite{funke2019using} & & \checkmark & 84.3 & 80.0\\
\hline
LC-SC-CRF~\cite{lea2016learning} & \checkmark & \checkmark & 83.5 & 76.8\\
Fusion-KV~\cite{qin2020temporal} & \checkmark & \checkmark & 86.3 & 87.2\\
\bfseries MRG-Net (Ours) & \checkmark & \checkmark &\bfseries 87.9 $\pm$ 4.2 &\bfseries 89.3 $\pm$ 5.2\\
\hline
\end{tabular}
}
\end{center}
\vspace{-4mm}
\end{table}

\begin{figure}[t]
    \centering
    \includegraphics[width=0.4\textwidth]{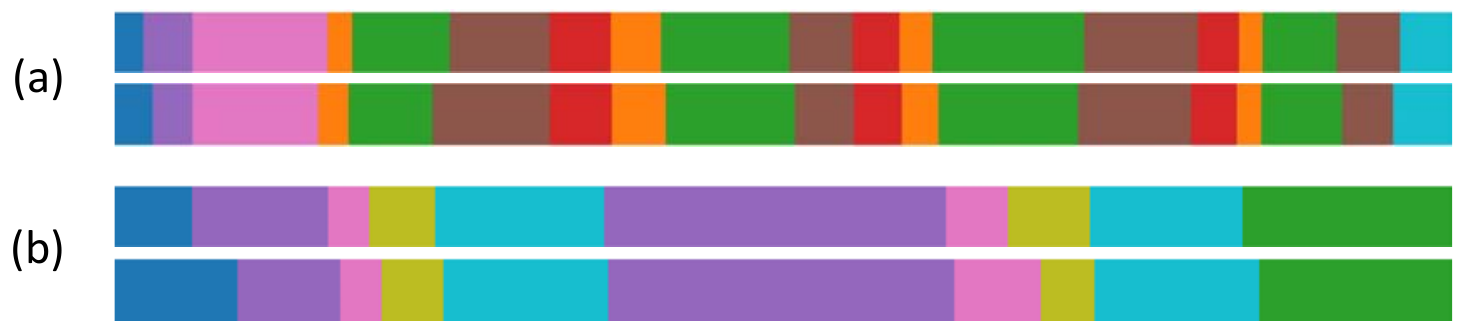}
    \vspace{-3mm}
    \caption{Color-coded ribbon illustration of surgical gesture recognition on Suturing task (a) and Knot Typing task (b) with ground truth (top) and our results (bottom).} \label{fig3}
    \vspace{-5mm}
\end{figure}

\begin{table}[t]
\begin{center}
\caption{Results of different methods on JIGSAWS Knot Typing dataset for gesture recognition.}\label{tab_knot}
\scalebox{1}{
\begin{tabular}{|c|c|c|c|c|}
\hline
\multirow{2}{*}{Methods} & \multicolumn{2}{|c|}{Input data} & \multirow{2}{*}{~Accuracy~}& \multirow{2}{*}{Edit Score} \\
\cline{2-3}
~ & Kin & Vid &&\\
\hline
SC-CRF~\cite{ahmidi2017dataset} & \checkmark &  & 78.9 & N/A \\
BoF~\cite{ahmidi2017dataset} & & \checkmark & 86.5 & N/A \\
MsM-CRF~\cite{ahmidi2017dataset} & \checkmark & \checkmark & 77.3 & N/A \\
\bfseries MRG-Net (Ours) & \checkmark & \checkmark &\bfseries 88.1 $\pm$ 3.8 &\bfseries 87.0 $\pm$ 6.8\\
\hline
\end{tabular}
}
\end{center}
\vspace{-4mm}
\end{table}

\begin{figure}[t]
\centering
\includegraphics[width=0.25\textwidth]{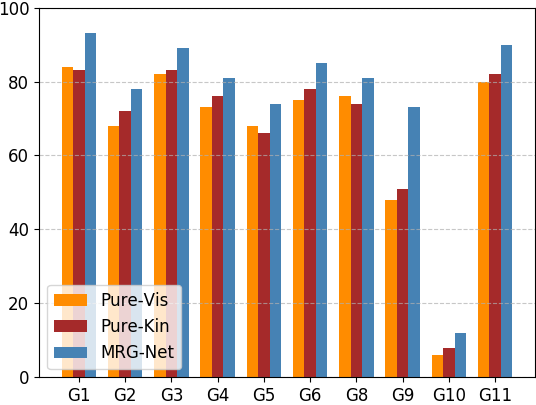}
\vspace{-3mm}
\caption{Bar chart of gesture-wise recognition accuracy.}
\label{fig2}
\vspace{-6mm}
\end{figure}

We compare our proposed MRG-Net with previous state-of-the-art methods on the benchmark dataset and we report the mean accuracy and mean Edit Score with standard deviation (std.) (those results without std. mean the original paper didn't report them).
 These methods are grouped into \emph{purely kinematics based} (i.e., SC-CRF~\cite{ahmidi2017dataset}, TCN~\cite{lea2016temporal}, Forward and Bidirectional LSTM~\cite{dipietro2016recognizing}, Bidirectional GRU~\cite{dipietro2019segmenting}, TricorNet~\cite{ding2017tricornet} of hybrid TCN and LSTM, APc~\cite{van2020multi} using multi-task RNN), \emph{purely video based} (i.e., TCN~\cite{lea2016temporal}, Policy+Value~\cite{gao2020automatic} of offline reinforcement learning, 3D CNN with post-processing~\cite{funke2019using}), and \emph{multi-modal based} methods (i.e., LC-SC-CRF~\cite{lea2016learning} and MsM-CRF~\cite{ahmidi2017dataset} with traditional machine learning, BoF~\cite{ahmidi2017dataset} with manual extracted features and Fusion-KV~\cite{qin2020temporal} with deep learning).

For the suturing task, which is the most popular task with more samples and gestures in JIGSAWS, we compare our results with eleven state-of-the-art methods listed in Table~\ref{tab1}.
We first see that our MRG-Net significantly outperforms the state-of-the-art uni-modal methods, with the accuracy exceeding the previous best kinematics based method~\cite{van2020multi} by 2.4\% and best video method~\cite{funke2019using} by 3.6\%. With multi-modal learning to capture the complementary information of visual and motion data, improved results are obtained, with the LC-SC-CRF~\cite{lea2016learning} outperforming six of the uni-modal methods on accuracy and the Fusion-KV~\cite{qin2020temporal} outperforming all of them. Importantly, our MRG-Net achieves the highest accuracy of 87.9\% (with lowest std. 4.2\%) and Edit Score of 89.3 (with lowest std. 5.2) compared among the multi-modal methods, demonstrating the superiority of our method enabling dynamic interactions for modeling the inherent relations of multiple input sources with graph learning.

For the knot typing task, which is more complex than suturing while less validated in previous works, we list the results in Table~\ref{tab_knot}. The performance of current state-of-the-art uni-modal and multi-modal methods are referenced from the benchmark in~\cite{ahmidi2017dataset}.
It can be observed that traditional multi-modal method (MsM-CRF),
if without sufficient integration of the visual and kinematics information in the complex task,
even obtained worse performance compared with pure video based method and pure kinematics based method.
Leveraging our proposed relational graph learning method to interact the visual and kinematics embeddings in the latent space, a high recognition accuracy of 88.1\% can be achieved on this task.

For qualitative results, Fig.~\ref{fig3} illustrates the visualization results on both suturing and knot typing tasks in the form of color-coded ribbon, demonstrating the temporal consistency and smoothness of the surgical gesture predictions leveraging the high-quality multi-modal representations.

\subsection{Ablation Analysis on Our Method}

\begin{table}[t]
\begin{center}
\caption{Ablation study on key components of our method using the same backbone on JIGSAWS Suturing dataset.}
\label{tab2}
\scalebox{1}{
\begin{tabular}{|c|c|c|c|c|}
\hline
\multirow{2}{*}{Methods} & \multicolumn{2}{|c|}{Input data} & \multirow{2}{*}{~Accuracy~}& \multirow{2}{*}{Edit Score} \\
\cline{2-3}
~ & Kin & Vid &&\\
\hline
Pure-Vis & & \checkmark & 81.7 $\pm$ 6.7 & 86.5 $\pm$ 6.9\\
Pure-Kin & \checkmark & & 82.6 $\pm$ 6.5 & 86.6 $\pm$ 7.5\\
TCN-KV (w/o split) & \checkmark & \checkmark & 86.1 $\pm$ 5.6 & 85.3 $\pm$ 7.1\\
TCN-KV & \checkmark & \checkmark & 86.2 $\pm$ 5.4 & 86.1 $\pm$ 6.4\\
GCN-KV & \checkmark & \checkmark & 86.8 $\pm$ 4.9 & 87.4 $\pm$ 6.5\\
\bfseries MRG-Net & \checkmark & \checkmark & \bfseries 87.9 $\pm$ 4.2 &  \bfseries 89.3 $\pm$ 5.2\\
\hline
\end{tabular}
}
\end{center}
\end{table}

\begin{figure}[t]
\centering
\vspace{-6mm}
\includegraphics[width=0.32\textwidth]{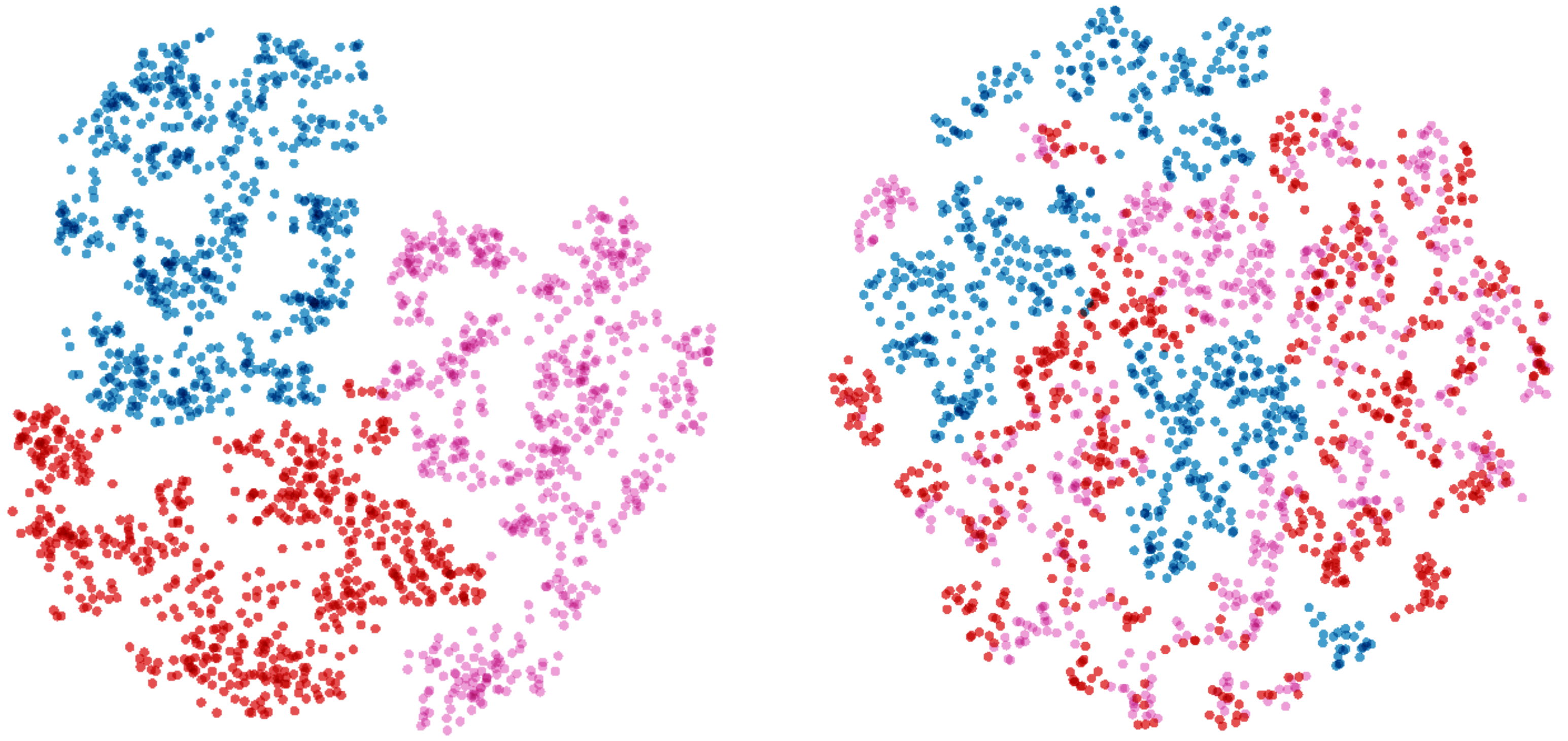}
\vspace{-2mm}
\caption{Embeddings of pre- (left) and post- (right) relational graph message propagation multi-modal features (blue: vision, red: left kinematics, pink: right kinematics). Best viewed in color.}
\label{fig4}
\vspace{-7mm}
\end{figure}

To validate the contribution of each key component in our proposed MRG-Net,
Table~\ref{tab2} lists the results of five ablation studies implemented with our own backbone on suturing task for direct comparison:
1) Pure-Vis: uni-modal using visual data,
2) Pure-Kin: uni-modal using kinematics,
3) TCN-KV (w/o split): merging video and kinematics (without splitting left and right arms) with TCN,
4) TCN-KV: merging video and kinematics (splitting left/right arms) with TCN,
5) GCN-KV: multi-modal learning with plain GCN without multi-relation,
and finally our proposed multi-relational MRG-Net.

We see that fusing visual and kinematics features in latent space can provide richer knowledge for achieving higher performance, even using simple concatenation of representations in temporal convolutional networks.
Note that splitting left and right kinematics data yields better results than treating both kinematics as a whole (comparing 3rd/4th rows with TCN), which also reveals the different information contained in the left and right ``hands" of robotic systems.
Moreover, our graph learning for interactive multi-modal message passing can bring improvement over TCN.
Further modelling multi-relations as designed based on domain knowledge, the gesture recognition performance gets higher, which confirms the significance of considering the ``edges" between ``nodes" diversely, as they incorporate distinct types of relations among various information sources.

In addition, we analyze the detailed accuracy across gesture category, as shown in Fig.~\ref{fig2}.
We notice a large variance in the results with the highest accuracy achieving 93\% (G1 ``reaching for needle with right hand"), while the lowest being less than 10\% (G10 ``loosening more suture").
The performance imbalance may be still due to the large variance in gesture frequency and sample numbers, which reflects the challenges in this recognition task which remains to be further conquered in future research.
Besides, we see that our relational graph multi-modal learning consistently outperforms Pure-Vis and Pure-Kin by a large margin (especially for G9 ``using right hand to help tighten suture" with strong visual/motion relationships), demonstrating the stable effectiveness of our method.
Last but not least, Fig.~\ref{fig4} visualizes the node features from MRG-Net learning process with t-SNE~\cite{maaten2008visualizing}, where the left and right embed the sets of $\{s_t,k_t^l,k_t^r\}$ and $\{\tilde{s}_t, \tilde{k}_t^l, \tilde{k}_t^r\}$, respectively, for observing feature clusters before and after multi-relational graph updates.
It clearly shows that multi-modal features are harmoniously fused from interactive message propagation and aggregation.

\begin{figure}[t]
  \centering
  \includegraphics[width=0.45\textwidth]{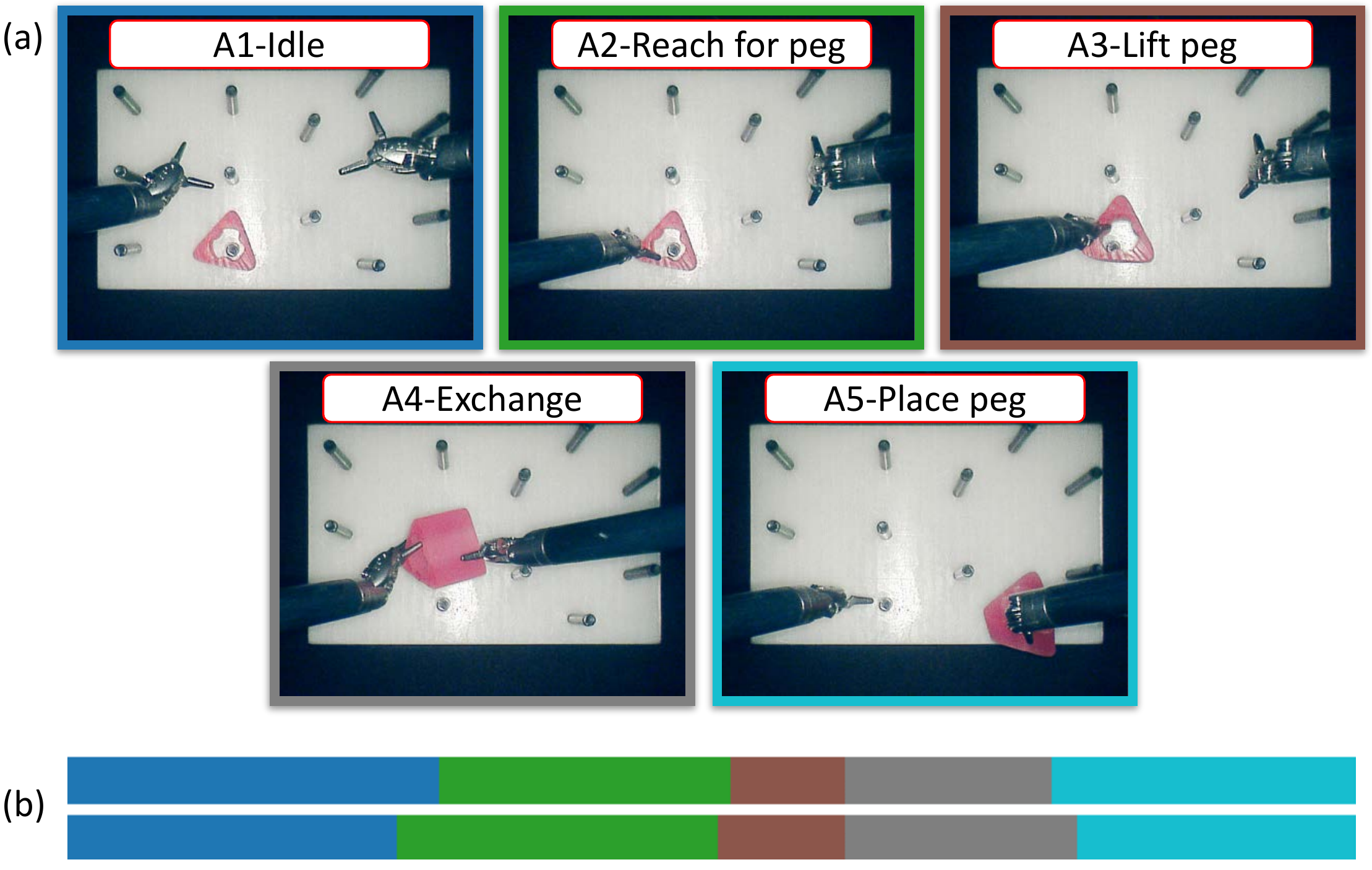}
  \vspace{-3mm}
  \caption{(a) Gestures of peg transfer, (b) Color-coded ribbon illustration of surgical gesture recognition on peg transfer task with ground truth (top) and our results (bottom).}
  \label{fig_peg}
  \vspace{-5mm}
\end{figure}

\subsection{Experiment on In-house dVRK Dataset from Two Centers}

To further validate our method, we have collected robotic multi-modal datasets on dVRK platforms in two centers from CUHK and JHU. The data collection conditions of the two datasets had variations in settings of
hand-eye calibrations, illuminations, operating locations, which reflected complications in real-world practice.
Both kinematics sequences and camera videos have been recorded and synchronized.

To build the in-house dVRK datasets, we experimented on the peg transfer task (see Fig.~\ref{fig_peg} (a)), which is one of the most popular tasks present in Fundamentals of Laparoscopic Surgery~\cite{ritter2007design} and widely adopted for surgical skill training~\cite{joseph2010chopstick}. Specifically, we defined and manually annotated five different gestures for peg transfer:
A1: Idle (No action performed);
A2: Reach for peg (with left hand);
A3: Lift peg (with left hand);
A4: Exchange (transfer the peg to right hand);
A5: Place peg (with right hand).
The dataset consists of 24 sequences with 12 sequences from CUHK and JHU each. The duration of sequences is within the range of 20-60 seconds due to different length settings to transfer the peg.
Within each site, all the operation records were performed by the same user
who is familiar with using dVRK platform. The kinematics data includes the position/orientation of end-effector and opening angle of the gripper, at the meanwhile, videos are synchronously recorded and there are all down-sampled to 10Hz in pre-processing.

Considering different conditions in data acquisition such as hand-eye settings of dVRK systems, appearances of peg transfer boards, we individually trained and tested models for each dataset, in which we split each dataset to perform 3-fold cross-validation (8 sequences for training and 4 for testing).
We adopted the same evaluation metrics as JIGSAWS (i.e., accuracy and Edit Score).
The results are listed in the Table~\ref{tab_peg_cuhk} and Table~\ref{tab_peg_jhu}, in which Baseline means the standard 2D CNN backbone used in MRG-Net (ResNet-18), while Pure-Vis and Pure-Kin represent the same configurations as Table~\ref{tab2}, which adopted TCN and LSTM.

It can be observed that, on both datasets, compared to 2D based Baseline, the methods of Pure-Vis and Pure-Kin can obtain higher accuracies and Edit Scores, leveraging their consideration of temporal information in the sequential data.
On both the datasets, our proposed MRG-Net achieves the highest accuracy and Edit Score, consistently outperforming the Baseline, Pure-Vis and Pure-Kin methods with a notable margin. The Edit Scores reaches as high as 98.7\% on CUHK dataset and 96.4\% on JHU dataset, which reflects the good smoothness and stability of the prediction on dVRK data. In addition, we notice that the model performances for CUHK dataset are overall slightly higher than that for JHU dataset.
We analyze that this is related to the variation of task duration between these two sites. The data recorded from JHU present larger diversity regarding task completion speed compared to CUHK data, thus may be more challenging for recognition. It will be interesting and valuable to further investigate model behavior differences between two datasets in our future work.

\begin{table}[t]
\begin{center}
\caption{Results of different methods on Peg Transfer dataset in site CUHK for gesture recognition.}\label{tab_peg_cuhk}
\scalebox{1}{
\begin{tabular}{|c|c|c|c|c|}
\hline
\multirow{2}{*}{Methods} & \multicolumn{2}{|c|}{Input data} & \multirow{2}{*}{~Accuracy~}& \multirow{2}{*}{Edit Score} \\
\cline{2-3}
~ & Kin & Vid &&\\
\hline
Baseline(ResNet-18) &  & \checkmark & 80.7 $\pm$ 7.4 & 35.1 $\pm$ 8.6\\
Pure-Vis &  & \checkmark & 88.9 $\pm$ 2.8 & 96.7 $\pm$ 3.9\\
Pure-Kin & \checkmark &  & 89.2 $\pm$ 2.5 & 95.6 $\pm$ 3.7\\
\bfseries MRG-Net (Ours) & \checkmark & \checkmark &\bfseries 91.0 $\pm$ 2.1 &\bfseries 98.7 $\pm$ 3.4\\
\hline
\end{tabular}
}
\end{center}
\vspace{-2mm}
\end{table}

\begin{table}[t]
\begin{center}
\caption{Results of different methods on Peg Transfer dataset in site JHU for gesture recognition.}\label{tab_peg_jhu}
\scalebox{1}{
\begin{tabular}{|c|c|c|c|c|}
\hline
\multirow{2}{*}{Methods} & \multicolumn{2}{|c|}{Input data} & \multirow{2}{*}{~Accuracy~}& \multirow{2}{*}{Edit Score} \\
\cline{2-3}
~ & Kin & Vid &&\\
\hline
Baseline(ResNet-18) &  & \checkmark & 78.5 $\pm$ 8.2 & 21.3 $\pm$ 9.8\\
Pure-Vis &  & \checkmark & 83.0 $\pm$ 3.6 & 95.1 $\pm$ 4.2\\
Pure-Kin & \checkmark &  & 85.1 $\pm$ 3.4 & 95.5 $\pm$ 3.9\\
\bfseries MRG-Net (Ours) & \checkmark & \checkmark &\bfseries 87.3 $\pm$ 2.9 &\bfseries 96.4 $\pm$ 3.6\\
\hline
\end{tabular}
}
\end{center}
\vspace{-7mm}
\end{table}

\section{Conclusion and Future Work}
\label{CONCLUSIONS}	
This paper presents a novel online multi-modal graph learning method to dynamically integrate complementary information in video and kinematics data from robotic systems, to achieve accurate surgical gesture recognition. Multi-relational representation aggregation is achieved through a designed directed graph to capture the underlying
joint knowledge
between the visual scenes and kinematics motions.
The effectiveness of our method is validated with state-of-the-art performance on the public dataset of JIGSAWS on two tasks of suturing and knot typing. Meanwhile, we investigate the significance of each component in our network by conducting ablation studies on JIGSAWS suturing dataset.
Furthermore, the proposed method is validated on our collected in-house dVRK datasets, shedding light on the general efficacy of our approach.


In our future work, we shall explore
how to resolve the data variance and domain gap due to different acquisition environments and hardware platforms of our two in-house datasets. Potentially, we will design a 6-DOF transformer (a trainable homogeneous mapping) to uniformly align kinematics data from different platforms to a common feature space, and rely on optical flow to tackle the visual gap among different environment. 
With the help of these methods, we can improve the generalization ability of our method, so as to make full use of more robotic surgery datasets and achieve a cross-platform training and testing scheme. Moreover, we will investigate how to extract the multi-modal embeddings with unsupervised learning schemes in order to reduce the annotation cost.
We will also apply the developed visual-kinematics based surgical gesture recognition to downstream scenarios such as sub-task automation for robotic surgery.

\newpage
\bibliographystyle{IEEEtran}
\bibliography{IEEEabrv,ref}

\begin{thebibliography}{10}
\providecommand{\url}[1]{#1}
\csname url@rmstyle\endcsname
\providecommand{\newblock}{\relax}
\providecommand{\bibinfo}[2]{#2}
\providecommand\BIBentrySTDinterwordspacing{\spaceskip=0pt\relax}
\providecommand\BIBentryALTinterwordstretchfactor{4}
\providecommand\BIBentryALTinterwordspacing{\spaceskip=\fontdimen2\font plus
\BIBentryALTinterwordstretchfactor\fontdimen3\font minus
  \fontdimen4\font\relax}
\providecommand\BIBforeignlanguage[2]{{%
\expandafter\ifx\csname l@#1\endcsname\relax
\typeout{** WARNING: IEEEtran.bst: No hyphenation pattern has been}%
\typeout{** loaded for the language `#1'. Using the pattern for}%
\typeout{** the default language instead.}%
\else
\language=\csname l@#1\endcsname
\fi
#2}}

\bibitem{moustris2011evolution}
G.~P. e.~a. Moustris, ``Evolution of autonomous and semi-autonomous robotic
  surgical systems: a review of the literature,'' \emph{The international
  journal of medical robotics and computer assisted surgery}, vol.~7, no.~4,
  pp. 375--392, 2011.

\bibitem{guthart2000intuitive}
G.~S. Guthart and J.~K. Salisbury, ``The intuitive/sup tm/telesurgery system:
  overview and application,'' in \emph{Proceedings 2000 ICRA. Millennium
  Conference. IEEE International Conference on Robotics and Automation.
  Symposia Proceedings (Cat. No. 00CH37065)}, vol.~1.\hskip 1em plus 0.5em
  minus 0.4em\relax IEEE, 2000, pp. 618--621.

\bibitem{maier2017surgical}
L.~Maier-Hein, S.~S. Vedula, S.~Speidel, N.~Navab, R.~Kikinis, A.~Park,
  M.~Eisenmann, H.~Feussner, G.~Forestier, S.~Giannarou, \emph{et~al.},
  ``Surgical data science for next-generation interventions,'' \emph{Nat.
  Biomed. Eng.}, vol.~1, no.~9, pp. 691--696, 2017.

\bibitem{poursartip2018analysis}
B.~Poursartip, M.-E. LeBel, R.~V. Patel, M.~D. Naish, and A.~L. Trejos,
  ``Analysis of energy-based metrics for laparoscopic skills assessment,''
  \emph{IEEE Transactions on Biomedical Engineering}, vol.~65, no.~7, pp.
  1532--1542, 2018.

\bibitem{nagy2019dvrk}
T.~D. Nagy and T.~Haidegger, ``A dvrk-based framework for surgical subtask
  automation,'' \emph{Acta Polytechnica Hungarica}, pp. 61--78, 2019.

\bibitem{tao2012sparse}
L.~Tao, E.~Elhamifar, S.~Khudanpur, G.~D. Hager, and R.~Vidal, ``Sparse hidden
  markov models for surgical gesture classification and skill evaluation,'' in
  \emph{IPCAI}.\hskip 1em plus 0.5em minus 0.4em\relax Springer, 2012, pp.
  167--177.

\bibitem{varadarajan2009data}
B.~Varadarajan, C.~Reiley, H.~Lin, S.~Khudanpur, and G.~Hager, ``Data-derived
  models for segmentation with application to surgical assessment and
  training,'' in \emph{International Conference on Medical Image Computing and
  Computer Assisted Intervention}.\hskip 1em plus 0.5em minus 0.4em\relax
  Springer, 2009, pp. 426--434.

\bibitem{zappella2013surgical}
L.~Zappella, B.~B{\'e}jar, G.~Hager, and R.~Vidal, ``Surgical gesture
  classification from video and kinematic data,'' \emph{Medical image
  analysis}, vol.~17, no.~7, pp. 732--745, 2013.

\bibitem{dipietro2016recognizing}
R.~DiPietro, C.~Lea, A.~Malpani, N.~Ahmidi, S.~S. Vedula, G.~I. Lee, M.~R. Lee,
  and G.~D. Hager, ``Recognizing surgical activities with recurrent neural
  networks,'' in \emph{International Conference on Medical Image Computing and
  Computer Assisted Intervention}.\hskip 1em plus 0.5em minus 0.4em\relax
  Springer, 2016, pp. 551--558.

\bibitem{van2020multi}
B.~Van~Amsterdam, M.~J. Clarkson, and D.~Stoyanov, ``Multi-task recurrent
  neural network for surgical gesture recognition and progress prediction,''
  \emph{arXiv preprint arXiv:2003.04772}, 2020.

\bibitem{gurcan2019surgical}
I.~Gurcan and H.~Van~Nguyen, ``Surgical activities recognition using
  multi-scale recurrent networks,'' in \emph{ICASSP 2019-2019 IEEE
  International Conference on Acoustics, Speech and Signal Processing
  (ICASSP)}.\hskip 1em plus 0.5em minus 0.4em\relax IEEE, 2019, pp. 2887--2891.

\bibitem{lea2017temporal}
C.~Lea, M.~D. Flynn, R.~Vidal, A.~Reiter, and G.~D. Hager, ``Temporal
  convolutional networks for action segmentation and detection,'' in
  \emph{proceedings of the IEEE Conference on Computer Vision and Pattern
  Recognition}, 2017, pp. 156--165.

\bibitem{jin2017sv}
Y.~Jin, Q.~Dou, H.~Chen, L.~Yu, J.~Qin, C.-W. Fu, and P.-A. Heng, ``Sv-rcnet:
  workflow recognition from surgical videos using recurrent convolutional
  network,'' \emph{IEEE transactions on medical imaging}, vol.~37, no.~5, pp.
  1114--1126, 2017.

\bibitem{funke2019using}
I.~Funke and et~al., ``Using 3d convolutional neural networks to learn
  spatiotemporal features for automatic surgical gesture recognition in
  video,'' in \emph{International Conference on Medical Image Computing and
  Computer Assisted Intervention}.\hskip 1em plus 0.5em minus 0.4em\relax
  Springer, 2019, pp. 467--475.

\bibitem{zhang2020symmetric}
J.~Zhang, Y.~Nie, Y.~Lyu, H.~Li, J.~Chang, X.~Yang, and J.~J. Zhang,
  ``Symmetric dilated convolution for surgical gesture recognition,'' in
  \emph{International Conference on Medical Image Computing and
  Computer-Assisted Intervention}.\hskip 1em plus 0.5em minus 0.4em\relax
  Springer, 2020, pp. 409--418.

\bibitem{murali2016tsc}
A.~Murali, A.~Garg, S.~Krishnan, F.~T. Pokorny, P.~Abbeel, T.~Darrell, and
  K.~Goldberg, ``Tsc-dl: Unsupervised trajectory segmentation of multi-modal
  surgical demonstrations with deep learning,'' in \emph{2016 IEEE
  International Conference on Robotics and Automation (ICRA)}.\hskip 1em plus
  0.5em minus 0.4em\relax IEEE, 2016, pp. 4150--4157.

\bibitem{zhao2018fast}
H.~Zhao, J.~Xie, Z.~Shao, Y.~Qu, Y.~Guan, and J.~Tan, ``A fast unsupervised
  approach for multi-modality surgical trajectory segmentation,'' \emph{IEEE
  Access}, vol.~6, pp. 56\,411--56\,422, 2018.

\bibitem{lea2016learning}
C.~Lea, R.~Vidal, and G.~D. Hager, ``Learning convolutional action primitives
  for fine-grained action recognition,'' in \emph{ICRA}.\hskip 1em plus 0.5em
  minus 0.4em\relax IEEE, 2016, pp. 1642--1649.

\bibitem{qin2020temporal}
Y.~Qin, S.~A. Pedram, S.~Feyzabadi, M.~Allan, A.~J. McLeod, J.~W. Burdick, and
  M.~Azizian, ``Temporal segmentation of surgical sub-tasks through deep
  learning with multiple data sources,'' \emph{arXiv preprint
  arXiv:2002.02921}, 2020.

\bibitem{qin2020davincinet}
Y.~Qin, S.~Feyzabadi, M.~Allan, J.~W. Burdick, and M.~Azizian, ``davincinet:
  Joint prediction of motion and surgical state in robot-assisted surgery,''
  \emph{arXiv preprint arXiv:2009.11937}, 2020.

\bibitem{dwivedi2020benchmarkgnns}
V.~P. Dwivedi, C.~K. Joshi, T.~Laurent, Y.~Bengio, and X.~Bresson,
  ``Benchmarking graph neural networks,'' \emph{arXiv preprint
  arXiv:2003.00982}, 2020.

\bibitem{scarselli2008graph}
F.~Scarselli, M.~Gori, A.~C. Tsoi, M.~Hagenbuchner, and G.~Monfardini, ``The
  graph neural network model,'' \emph{IEEE Transactions on Neural Networks},
  vol.~20, no.~1, pp. 61--80, 2008.

\bibitem{zhou2018graph}
J.~Zhou, G.~Cui, Z.~Zhang, C.~Yang, Z.~Liu, L.~Wang, C.~Li, and M.~Sun, ``Graph
  neural networks: A review of methods and applications,'' \emph{arXiv preprint
  arXiv:1812.08434}, 2018.

\bibitem{kipf2016semi}
T.~N. Kipf and M.~Welling, ``Semi-supervised classification with graph
  convolutional networks,'' \emph{arXiv preprint arXiv:1609.02907}, 2016.

\bibitem{wang2018zero}
X.~Wang, Y.~Ye, and A.~Gupta, ``Zero-shot recognition via semantic embeddings
  and knowledge graphs,'' in \emph{CVPR}, 2018, pp. 6857--6866.

\bibitem{marcheggiani2018exploiting}
D.~Marcheggiani, J.~Bastings, and I.~Titov, ``Exploiting semantics in neural
  machine translation with graph convolutional networks,'' \emph{arXiv
  preprint:1804.08313}, 2018.

\bibitem{wang2018deep}
Z.~Wang, T.~Chen, J.~Ren, W.~Yu, H.~Cheng, and L.~Lin, ``Deep reasoning with
  knowledge graph for social relationship understanding,'' \emph{arXiv preprint
  arXiv:1807.00504}, 2018.

\bibitem{wang2019graph}
S.~Wang, Z.~Xu, C.~Yan, and J.~Huang, ``Graph convolutional nets for tool
  presence detection in surgical videos,'' in \emph{IPMI}.\hskip 1em plus 0.5em
  minus 0.4em\relax Springer, 2019, pp. 467--478.

\bibitem{weibel2019robust}
J.-B. Weibel, T.~Patten, and M.~Vincze, ``Robust 3d object classification by
  combining point pair features and graph convolution,'' in \emph{2019
  International Conference on Robotics and Automation (ICRA)}.\hskip 1em plus
  0.5em minus 0.4em\relax IEEE, 2019, pp. 7262--7268.

\bibitem{weibel2019addressing}
J.~B. Weibel, T.~Patten, and M.~Vincze, ``Addressing the sim2real gap in
  robotic 3-d object classification,'' \emph{IEEE Robotics and Automation
  Letters}, vol.~5, no.~2, pp. 407--413, 2019.

\bibitem{sarikaya2020towards}
D.~Sarikaya and P.~Jannin, ``Towards generalizable surgical activity
  recognition using spatial temporal graph convolutional networks,''
  \emph{arXiv preprint:2001.03728}, 2020.

\bibitem{he2016deep}
K.~He, X.~Zhang, S.~Ren, and J.~Sun, ``Deep residual learning for image
  recognition,'' in \emph{Proceedings of the IEEE conference on computer vision
  and pattern recognition}, 2016, pp. 770--778.

\bibitem{gilmer2017neural}
J.~Gilmer, S.~S. Schoenholz, P.~F. Riley, O.~Vinyals, and G.~E. Dahl, ``Neural
  message passing for quantum chemistry,'' in \emph{Proceedings of the 34th
  International Conference on Machine Learning-Volume 70}.\hskip 1em plus 0.5em
  minus 0.4em\relax JMLR. org, 2017, pp. 1263--1272.

\bibitem{tsai1989new}
R.~Y. Tsai, R.~K. Lenz, \emph{et~al.}, ``A new technique for fully autonomous
  and efficient 3 d robotics hand/eye calibration,'' \emph{IEEE Transactions on
  robotics and automation}, vol.~5, no.~3, pp. 345--358, 1989.

\bibitem{li2018deeper}
Q.~Li, Z.~Han, and X.-M. Wu, ``Deeper insights into graph convolutional
  networks for semi-supervised learning,'' in \emph{AAAI}, 2018.

\bibitem{schlichtkrull2018modeling}
M.~Schlichtkrull, T.~N. Kipf, P.~Bloem, R.~Van Den~Berg, I.~Titov, and
  M.~Welling, ``Modeling relational data with graph convolutional networks,''
  in \emph{ESWC}.\hskip 1em plus 0.5em minus 0.4em\relax Springer, 2018, pp.
  593--607.

\bibitem{wang2019dgl}
M.~Wang, L.~Yu, D.~Zheng, Q.~Gan, Y.~Gai, Z.~Ye, M.~Li, J.~Zhou, Q.~Huang,
  C.~Ma, \emph{et~al.}, ``Deep graph library: Towards efficient and scalable
  deep learning on graphs,'' \emph{arXiv preprint arXiv:1909.01315}, 2019.

\bibitem{kingma2014adam}
D.~P. Kingma and J.~Ba, ``Adam: A method for stochastic optimization,''
  \emph{arXiv preprint arXiv:1412.6980}, 2014.

\bibitem{gao2014jhu}
Y.~Gao, S.~S. Vedula, C.~E. Reiley, N.~Ahmidi, B.~Varadarajan, H.~C. Lin,
  L.~Tao, L.~Zappella, B.~B{\'e}jar, D.~D. Yuh, \emph{et~al.}, ``Jhu-isi
  gesture and skill assessment working set (jigsaws): A surgical activity
  dataset for human motion modeling,'' in \emph{MICCAI Workshop: M2CAI},
  vol.~3, 2014, p.~3.

\bibitem{ahmidi2017dataset}
N.~Ahmidi, L.~Tao, S.~Sefati, Y.~Gao, C.~Lea, B.~B. Haro, L.~Zappella,
  S.~Khudanpur, R.~Vidal, and G.~D. Hager, ``A dataset and benchmarks for
  segmentation and recognition of gestures in robotic surgery,'' \emph{IEEE
  Transactions on Biomedical Engineering}, vol.~64, no.~9, pp. 2025--2041,
  2017.

\bibitem{lea2016segmental}
C.~Lea, A.~Reiter, R.~Vidal, and G.~D. Hager, ``Segmental spatiotemporal cnns
  for fine-grained action segmentation,'' in \emph{ECCV}.\hskip 1em plus 0.5em
  minus 0.4em\relax Springer, 2016, pp. 36--52.

\bibitem{lea2016temporal}
C.~Lea, R.~Vidal, A.~Reiter, and G.~D. Hager, ``Temporal convolutional
  networks: A unified approach to action segmentation,'' in \emph{ECCV}.\hskip
  1em plus 0.5em minus 0.4em\relax Springer, 2016, pp. 47--54.

\bibitem{ding2017tricornet}
L.~Ding and C.~Xu, ``Tricornet: A hybrid temporal convolutional and recurrent
  network for video action segmentation,'' \emph{arXiv preprint
  arXiv:1705.07818}, 2017.

\bibitem{dipietro2019segmenting}
R.~DiPietro, N.~Ahmidi, A.~Malpani, M.~Waldram, G.~I. Lee, M.~R. Lee, S.~S.
  Vedula, and G.~D. Hager, ``Segmenting and classifying activities in
  robot-assisted surgery with recurrent neural networks,'' \emph{IJCARS},
  vol.~14, no.~11, pp. 2005--2020, 2019.

\bibitem{gao2020automatic}
X.~Gao, Y.~Jin, Q.~Dou, and P.-A. Heng, ``Automatic gesture recognition in
  robot-assisted surgery with reinforcement learning and tree search,''
  \emph{arXiv preprint arXiv:2002.08718}, 2020.

\bibitem{maaten2008visualizing}
L.~v.~d. Maaten and G.~Hinton, ``Visualizing data using t-sne,'' \emph{Journal
  of machine learning research}, vol.~9, no. Nov, pp. 2579--2605, 2008.

\bibitem{ritter2007design}
E.~M. Ritter and D.~J. Scott, ``Design of a proficiency-based skills training
  curriculum for the fundamentals of laparoscopic surgery,'' \emph{Surgical
  innovation}, vol.~14, no.~2, pp. 107--112, 2007.

\bibitem{joseph2010chopstick}
R.~A. Joseph, A.~C. Goh, S.~P. Cuevas, M.~A. Donovan, M.~G. Kauffman, N.~A.
  Salas, B.~Miles, B.~L. Bass, and B.~J. Dunkin, ``“chopstick” surgery: a
  novel technique improves surgeon performance and eliminates arm collision in
  robotic single-incision laparoscopic surgery,'' \emph{Surgical endoscopy},
  vol.~24, no.~6, pp. 1331--1335, 2010.

\end{thebibliography}

\end{document}